**Enhancing Diameter Measurement Accuracy in Machine Vision Applications**

Ahmet Gökhan POYRAZ[1,2], Ahmet Emir DİRİK[3], Hakan GÜRKAN[1], Mehmet KAÇMAZ[4]

Department of Electrical and Electronics Engineering, Bursa Technical University, 16310 Bursa, Turkey[1]

Doğu Pres R&D, 1610, Bursa, Turkey[2]

Department of Computer Engineering, Bursa Uludağ University, 16120 Bursa, Turkey[3]

Institute of Electrical Information Technology, Clausthal University of Technology, Clausthal-Zellerfeld, Germany[4]

Author's e-mail: agpoyraz@gmail.com (Corresponding Author), edirik@uludag.edu.tr, hakan.gurkan@btu.edu.tr

**Abstract**

In camera measurement systems, specialized equipment such as telecentric lenses is often employed to measure parts with narrow tolerances. However, despite the use of such equipment, measurement errors can occur due to mechanical and software-related factors within the system. These errors are particularly evident in applications where parts of different diameters are measured using the same setup. This study proposes two innovative approaches to enhance measurement accuracy using multiple known reference parts: a conversion factor-based method and a pixel-based method. In the first approach, the conversion factor is estimated from known references to calculate the diameter (mm) of the unknown part. In the second approach, the diameter (mm) is directly estimated using pixel-based diameter information from the references. The experimental setup includes an industrial-grade camera and telecentric lenses. Tests conducted on glass samples (1–12 mm) and metal workpieces (3–24 mm) show that measurement errors, which originally ranged from 13–114 μm, were reduced to 1–2 μm using the proposed methods. By utilizing only a few known reference parts, the proposed approach enables high-accuracy measurement of all parts within the camera's field of view. Additionally, this method enhances the existing diameter measurement literature by significantly reducing error rates and improving measurement reliability.

1. **Introduction**

With the rapid advancement of technology, machine vision applications have become an integral part of various industries, including automotive, aerospace, food, and lumber. In the automotive sector, the demand for speed and precision has made camera-based inspection systems indispensable. The literature provides extensive studies on machine vision [1]. For instance, some studies focus on automated quality control systems for assessing the quality of fruits and vegetables [2,3], while others explore classification applications using robotic arms under varying lighting conditions [4]. In addition to visually segregating inspected parts, camera-based measurements are conducted in many applications. Many researchers have carried out measurement applications oriented towards industry. Various methods for shaft and gear measurement have been developed by researchers. Gadelmawla [10] developed a program for gear measurement and stated that it achieves the desired measurements with high precision. Duan et al. [11] proposed a high-precision edge-finding method for measuring mechanical parts. Du et al. [12] proposed a gear measurement method using machine vision and suggested a measurement methodology that can measure all the features of the gear. In [13], the shaft size was measured using a lighting system. Sub-pixel-based approaches have been utilized in some studies for length measurement. [14] measured mini milling cutters using telecentric lenses and backlighting, adopting grayscale histogram and threshold segmentation methods to reach sub-pixel level. They achieved measurements with an uncertainty of approximately 1 μm using their developed algorithm and mechanical system. Chen et al. [15] conducted dimensional measurements at sub-pixel level using logistic edge models. [16] proposed a line measurement algorithm for machine vision applications. Wang et al. [17] developed a sub-pixel-based measurement system for measuring printed circuit boards. [18, 19, 20, 21] studies have used machine vision-based systems to measure the dimensions of parts.

In industrial camera-based measurement applications, custom reference gauges are specifically manufactured to convert pixel measurements to actual dimensions (mm). These references are produced according to the dimensions of the workpiece and their actual values are determined using specialized equipment, such as a Coordinate Measurement Machine (CMM). After the reference part is measured in pixel units on the current system, a conversion factor is calculated. This conversion factor is then used to estimate the dimensions of other workpieces. The literature review indicates that the mm/px value varies depending on the measurements. Therefore, as the dimensions of the workpiece in serial control deviate further from those of the reference part, measurement errors occur.

In industrial applications, it is often desired to measure multiple types of workpieces, rather than a single type, using the same camera-based system. At this point, if a single conversion factor is employed, errors, as highlighted in the literature, become inevitable. Practitioners address this issue by producing dedicated reference gauges for each type of workpiece. However, this approach incurs temporal and financial costs due to the production of multiple reference parts. Furthermore, when a workpiece of an unknown approximate size is introduced to the same camera-based system, the absence of a suitable conversion factor necessitates either initiating the reference part production process anew or proceeding with the known conversion factor, disregarding potential errors.

The motivation for this study stems from the need for precise measurement of the diameters of parts with different dimensions using the same mechanism, while avoiding the high costs associated with producing dedicated reference gauges. To address these issues, this paper proposes a generalizable approach that operates independently of specific measurement algorithms. The contributions of this study are outlined as follows:

1. Reducing the error arising from using a single conversion factor ($R$).

2. Preventing the need for separate reference gauges for each type, thus achieving significant time and cost savings.

3. Enabling high-precision measurement of workpieces with unknown approximate dimensions using only a limited number of reference gauges, without requiring additional gauge production.

The proposed approach utilizes a limited number of reference parts to accurately estimate both the appropriate conversion factor and the approximate dimensions of the workpiece. Two methods are introduced: conversion factor-based and pixel-based, both designed to reduce errors caused by using a single conversion factor. Experimental results validate the effectiveness of the approach, demonstrating its ability to eliminate the need for multiple reference parts while ensuring high-precision measurements and significant cost savings.

The innovative aspect of this study lies in its ability to overcome the limitations of traditional diameter measurement methods by introducing a universally applicable approach that operates independently of specific algorithms. Unlike conventional techniques, which require dedicated reference parts for each workpiece type, the proposed method achieves high-precision measurements using only a few reference samples. This approach significantly reduces the time and cost associated with producing reference gauges. Furthermore, the study examines and models the variation in mm/px conversion factors across different diameters, enabling accurate predictions for parts with unknown sizes. The integration of sub-pixel-based measurement algorithms further enhances the precision and reliability of the method. By addressing the gap between existing techniques and industrial requirements, the study provides a versatile and efficient solution for camera-based measurement applications.

2. **Related Work**

In this study, an in-depth literature review was conducted to enhance the understanding of the proposed approach and to better highlight its innovative aspects. The review focused on three main topics: measurement based on sub-pixel level, conversion factor, and error correction. Each topic is discussed in separate paragraphs.

To ensure precise and accurate measurements, both hardware and software must be of a high standard. Usually, edge points on the image of the component that needs to be measured are found before measurement. Then, the measurement result is calculated by combining these edge points. In addition to techniques that locate points at the pixel level, such as Sobel, Robert, and Canny algorithms, techniques operating at the sub-pixel level are crucial for measurement precision. At the sub-pixel level, numerous research studies are available. Tichem and Cohen [5] introduced the sub-pixel approach by examining pixel positions. Steger [6] proposed a method at the sub-pixel level for line and edge extraction. Trujillo-Pino et al. [7] proposed a regional area-based edge detection algorithm. In their proposed method, they accurately detected edge points using windows of certain sizes at transition regions of the image. Von Gioi and Randall [8] developed a sub-pixel edge detection algorithm by incorporating and enhancing the Devernay and Canny algorithms. Wang et al. [9] worked on a sub-pixel test pattern using Canny and Zernike moments. After detecting edge regions, the desired measurement is estimated by fitting with the found points. Seo et al. [22] proposed a method based on the summation of normalized derivatives to determine the sub-pixel position. Fabijanska [23] utilized the properties of the Gaussian function to find the edges of images containing noise at the sub-pixel level. Breder et al. [24] performed edge detection at the sub-pixel level by adopting the general B-spline model. Li et al. [25] proposed two bilinear interpolation-based sub-pixel measurement algorithms to perform small part measurements. Haibing e al. [26] proposed a cubic spline interpolation-based sub-pixel measurement algorithm for the diameter measurement of ring-type components. Ye et al. [27] proposed a method capable of detecting edges with high sharpness at the sub-pixel level. Zhao et al. [28] utilized the Facet model to determine sub-pixel positions. Sun et al. [29] employed the arctangent edge model for sub-pixel edge detection. Xie [30] utilized Zernike moments for industrial sub-pixel edge detection. Hagara et al. [31] developed a sub-pixel based fitting algorithm intended for industrial measurements. Xu et al.[34] proposed a sub-pixel position determination method for performing measurements on images. Zhang et al. [35] proposed a second derivative and curve fitting-based sub-pixel edge detection method. Zixin [36] utilized the Otsu thresholding method before the Zernike moment technique to quickly detect sub-pixels in boundary extraction. Mai et al. [37] employed the Zernike moment with a 9×9 mask for edge detection. [32] used a connected component-based method for industrial image processing applications. They count the pixels in the binary image to calculate the area. Recent research [33] proposed a sub-pixel counting-based method in their study. In their proposed method, they obtained precise measurements by considering gray pixels at the edge differently while calculating the area of the workpiece image. They also provided a comprehensive literature review.

Regardless of the algorithm used in measurement methods, conversion factor is employed when transitioning from pixel measurements to real-world units. In many different research studies, a conversion factor (also known as scale factor, reference unit pixel, convertion ratio, mm/px etc.) has been utilized. Su et al. [38] presents a machine vision system that is intended to provide a novel technique for detecting microdrill flank wear in order to meet the need for examining the relationship between drilling parameters and tool longevity. They utilized a conversion factor to transition to actual measurements. Busca et al. [39] utilized a conversion factor in their image processing algorithm used to measure vibrations on railway bridges, obtaining results in real measurement units. Phansak et al. [40] used mm/px value to measure part in a flexible automatic assembly system. Che et al. [41] utilize a conversion factor in their developed system to measure the diameter of the workpiece without stopping the turning process. Paul et al. [42] utilized reflection properties to measure the liquid tank's fill level. They calculated the mm/px value to obtain the actual measurement. In the field of structural health monitoring, Zhao et al. [43] used scale transformation factor in their studies. Dinh Do Van [44] proposed a Faster R-CNN-based image processing algorithm for defect detection in mechanical parts. To convert the algorithm's output to real units, a scale factor (mm/px) was utilized. In a study in the field of agriculture Christina et al. [45] proposed a specific rational model by taking into account the seed area measured at various spatial scales in relation to the pixel count using conversion ratio.

As can be understood from the literature, conversion factors are used in many applications. However, measurement deviations can occur due to the conversion factor. For this reason, researchers have calculated different conversion factors for each different measurement. In their studies [46,47,48], they have calculated a specific conversion factor for all types of rectangular parts in order to perform precise measurements. They have also examined the variations in the calculated mm/px values. [49] has calculated different conversion factors for the inner and outer diameters in the machine vision system they established to measure the diameters and concentricity values of ring-type parts. They have shown that the conversion factors for the inner and outer diameters are different from each other. [50] has used the Simpson algorithm to measure the part diameter. They have clearly shown that the error increases as the calibration diameter deviates from the measurement diameter. To solve this problem, they have used multiple conversion factors by using different parts. In the literature, there are various studies to reduce the error caused by the variability in mm/px. [51] has corrected the error in distance measurement. They have stated that the mm/px value changes as the distance changes. They have claimed to reduce the error with their proposed error correction model. In their study [52], they modeled and estimated the measurement error caused by the height difference when measuring the diameters of cylinders with different diameters. [53] has calculated the mm/px values based on the distance of the part to the camera in order to measure the diameter of a cylindrical part in their camera system. Thus, they have created an error correction model that can predict the expected diameter errors. In certain studies, researchers have proposed their own error correction models for application-specific needs. [54] has presented an algorithm for measuring the age of trees by measuring the rings in cross-sections. They have also created an error correction model using least squared polynomial fitting to prevent geometric distortion. [55] has used Neural Network (NN) for accurate measurement in computer vision applications. In their method, they have proposed an NN-based error correction model to correct the error.

### 3. Measurement Error Analysis
*3.1 Problem Definition and Measurement Error Analysis*

The measurement algorithms in the literature have achieved significant success due to their robust approaches. However, despite their effectiveness, measurement errors persist. These errors can be attributed to two primary factors: mechanical and software-related. Mechanical factors include lens properties, camera sensor specifications, light transmission through the measurement glass, lighting conditions, alignment of the glass, focus distortions, and external noise. On the software side, the performance of the measurement algorithm plays a critical role in determining the accuracy and precision of measurements. As the diameter being measured decreases and tolerances become narrower, the importance of algorithm performance increases significantly.

Using the same $R$ value for measuring workpieces with varying diameters is not ideal for precision measurement applications. This is because keeping the $R$ value constant leads to errors as the measured diameter diverges from the reference diameter. These errors become more pronounced as the measured diameter deviates further from the diameter used to calculate the $R$ value. Studies in the literature have highlighted this issue [50, 51, 52, 53]. To better illustrate this problem and clearly define the issue, a test was conducted. A dataset was generated using glass reference gauges with diameters of 1, 1.5, 2, 3, 4, 5, 6, 8, and 12 mm. Images were captured for each diameter, and measurements in pixel units were calculated using three different algorithms. To minimize repeatability errors, the average of multiple measurements was taken for each diameter. The $R$ value was calculated based on a reference diameter of 5 mm, and this value was subsequently used to convert pixel measurements to millimeters for the other diameters. Measurement errors were then computed by comparing the measured values with the actual dimensions. The results are presented in Figure 1.

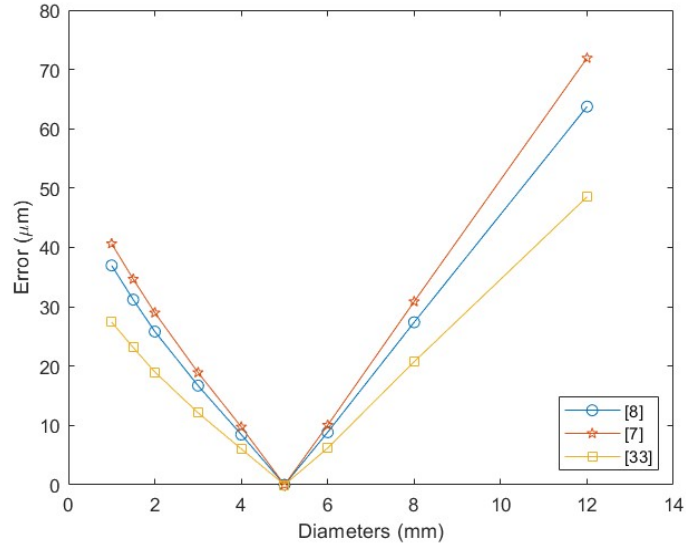

**Figure 1.** Errors corresponding to diameters when only single R value(for 5 mm) is used

The primary problem addressed in this paper arises from measurement errors caused by both the software and mechanical setup. As shown in Figure 1, using a single reference gauge results in similar errors across all three algorithms. To mitigate this issue, practitioners often use a separate reference gauge for each diameter, enabling a unique $R$ value for each (Figure 2). However, this approach incurs significant financial and production costs. Moreover, when an intermediate diameter needs to be measured, a new reference gauge becomes necessary. This challenge forms the motivation behind our study: eliminating the measurement errors caused by the system. To address this, we propose conversion factor and pixel-based predictive models as a solution.

*3.2 Relationship Between Diameter and Conversion Factor*

The conversion factor ($R$) is dependent on two variables as defined in Equation 1: the actual diameter of the reference part and the corresponding pixel-based measurement. Since reference parts are typically custom-made with low uncertainties, the $R$ value primarily depends on the measurement value in pixel units. However, mechanical and software factors influence the pixel-based measurement, introducing variability in $R$. Consequently, the error varies with the diameter, as depicted in the V-shaped Error-Diameter graph (Figure 1) or the $R$-Diameter curve (Figure 2).

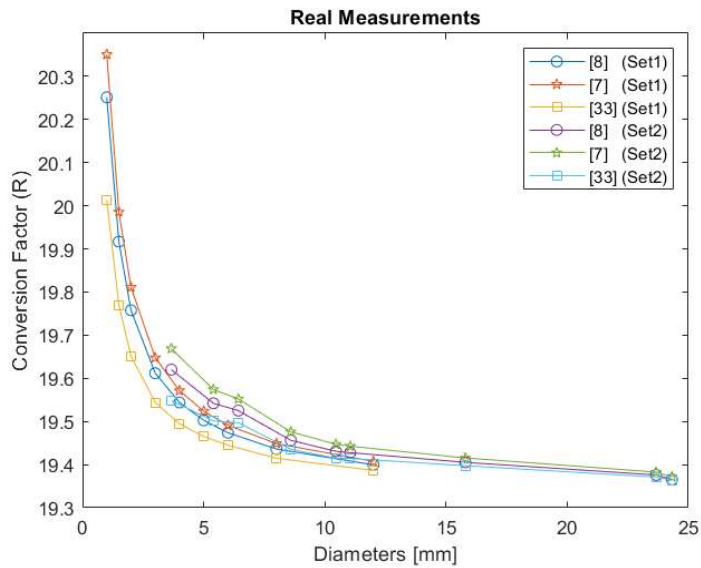

**Figure 2.** R-Diameter graph derived from the Set1 (glass) and Set2 (metal) datasets.

In Figure 2, the $R$ values obtained from the real experimental setup are plotted against the diameters. The experimental data, detailed in the experimental setup section, include glass reference gauges (diameter range: 1–12 mm) and metal reference gauges (diameter range: 3–24 mm). Examination of Figure 2 reveals that $R$ values vary with diameter, with smaller diameters yielding higher $R$ values, while larger diameters result in lower $R$ values. This trend is consistent across both data sets. Additionally, the graph demonstrates a convergence behavior of the curves toward a specific value, independent of the measurement algorithms or data sets. The observed similarities between the graphs suggest that R-related variability can be predicted using the proposed methods. These findings form the foundation for the R-based and pixel-based diameter estimation approaches presented in this study.

## 4. Proposed Approach
### 4.1 Conversion Factor

In camera measurement systems, calculating the desired measurements of workpieces requires first determining the measurement algorithm. Once the algorithm is defined, the captured images are processed to calculate the desired measurements in pixel units. Since these systems calculate values in pixel units, they must be converted to real-world units. In other words, it is necessary to determine how much real-world distance corresponds to one pixel (Figure 1). To determine this, the conversion factor $R$ is calculated using the formula:

$$R = \frac{Known\ Diameter\ in\ mm\ (D)}{Measured\ Diameter\ in\ pixel\ (P)} \tag{1}$$

To calculate $R$, a reference gauge with a known dimension is required. These gauges are typically made from materials such as ceramic, glass, or metal and are specially manufactured. Alternatively, a standard workpiece can be measured using a CMM to establish the reference. However, in this case, uncertainties from both the workpiece and the CMM can negatively affect the $R$ value. For this reason, the first approach is generally preferred. Assuming the reference gauge error is negligible, the $R$ value will primarily depend on the measured diameter. Since this study focuses on diameter measurements, "measurement" will refer specifically to "diameter" throughout the rest of the paper. To convert the diameter value obtained in pixel units to real-world units, the pixel-based diameter is multiplied by the $R$ value. The result is the diameter expressed in millimeters (mm).

In a system with $n$ reference parts, the diameter values of the reference parts can be expressed as $D = [D_1, D_2, \dots, D_n]$, their pixel-based measurement values as $P = [P_1, P_2, \dots, P_n]$ and the conversion factors obtained using Equation 1 as $R = [R_1, R_2, \dots, R_n]$. Here, the subscript $i$ indicates the index of the reference part. For each of the $n$ parts, the conversion factor $R_i$ is calculated as $R_i = D_i/P_i$. When a query part with an unknown diameter arrives, the estimated diameter in millimeters ($D_{est}$), is calculated based on the query part's pixel-based diameter ($P_q$) using Equation 2:

$$D_{est} = R_i \cdot P_q \tag{2}$$

If an approximate diameter of the query part is known, the $R_i$ value of the reference part closest in diameter can be used. If the approximate diameter is not known, $R_i$ value (where $i$ ranges from 1 to $n$) or average $R$ can be selected. The revised version of the equation for index $i$ is given as:

$$D_{est} = \frac{D_i}{P_i} \cdot P_q \tag{3}$$

To quantify the error between the measurement and the actual value, the error $E$ is defined in Equation 4, where $D_q$ represents the actual diameter of the query part:

$$E = \left| D_q - \frac{D_i}{P_i} \times P_q \right| \tag{4}$$

### 4.2 Proposed Estimation Methods:

To address the mentioned problem, two different methods are proposed: conversion factor-based and pixel-based approaches. Both methods aim to estimate the diameter of the query workpiece in millimeters. The proposed approach relies on using a few reference workpieces with known diameters to predict the millimeter-scale diameter of unknown workpieces. In the first method, the conversion factor-based approach, a model is created by considering the variation of the $R$ value with respect to different diameters. In the second method, the pixel-based approach, the diameter in millimeters is directly estimated using the pixel information of the measured diameter. Both approaches rely on known reference values to construct the models, which are then used to predict the diameters of query workpieces. To provide a clearer understanding of the application of these approaches, a concise explanation is presented in Table 1 and Figure 3.

**Table 1:** Proposed approaches

| Title | Method | Explanation | Approach |
|---|---|---|---|
| R-Based | M0 (Base) | Getting average R value and estimate measurements | Conversion Factor Based Approaches (px → R → mm) |
| | M1 | Estimate $R_{est}$ using nearest $R$ value from references | |
| | M2 | Estimate $R_{est}$ using nearest two $R$ values from references | |
| P-Based | M3 | Estimate $D_{est}$ using nearest two P values | Diameter Based Approaches (px → mm) |
| | M4 | Estimate $D_{est}$ using all P values | |

In the first approach, an $R$-based method is proposed, which considers the variation of the $R$ value with respect to different diameters. The objective is to predict the estimated conversion factor ($R_{est}$) and then calculate the actual diameter ($D_{est}$). Essentially, the known conversion factor values ($R$) and their corresponding diameters in pixels ($P$) are used to estimate the $R_{est}$ value for the query image (Figure 3a). Once the $R_{est}$ is determined, the actual diameter ($D_{est}$) of the query workpiece can be estimated using its measured diameter in pixels ($P_q$). This approach assumes the availability of a few (four in this paper) known reference values, which are used to construct the model and perform the estimation. In the second approach, the diameters of the reference parts in millimeters ($D$) and their corresponding pixel-based measurements ($P$) are directly utilized to predict the diameter ($D_{est}$) of the query workpiece (Figure 3b). Since the diameter is estimated directly in this method, no additional processing, such as calculating conversion factors, is required.

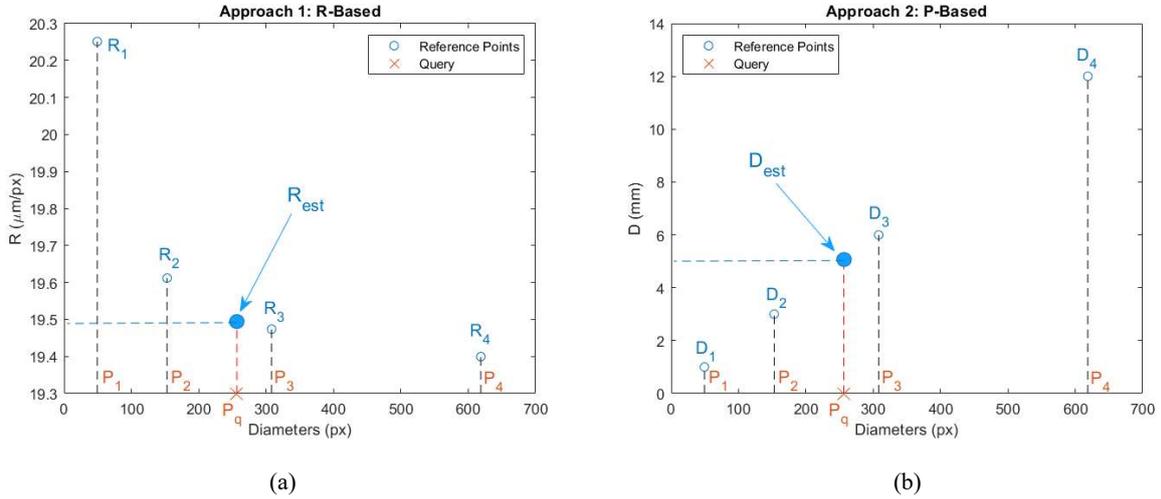

(a)  (b)

**Figure 3.** Proposed approaches estimation plots using reference $R$ and $D$ values. a. Estimation visualization of query part for approach-1. b. Estimation visualization of query part for approach-2.

The proposed methods estimate the diameter ($D_{est}$) of a query part based on its pixel-based measurement ($P_q$), using reference data including known diameters ($D$), pixel-based measurements of reference parts ($P$), and their conversion factors ($R$). The following steps outline the process:

Input:

- $\mathbf{D} = \{D_1, D_2, \ldots, D_n\}$: Known diameters (in mm) of reference parts.

- $\mathbf{P} = \{P_1, P_2, \ldots, P_n\}$: Pixel-based diameter measurements of reference parts.

- $\mathbf{R} = \{R_1, R_2, \ldots, R_n\}$: Conversion factors for reference parts.

- $P_q$: Pixel-based measurement of the query part.

- $method$: Chosen method for estimation.

Output:

- $D_{est}$: Estimated diameter (in mm).

### 4.2.1 R-Based Estimation Methods:

*Method 0 (base)*: In this method, the estimated diameter is calculated using the average of all conversion factors ($R$) obtained from the reference parts. The average conversion factor ($R_{est}$) is computed as:

$$R_{est} = \frac{1}{n} \sum_{i=1}^{n} R_i$$

Using this average, the estimated diameter $D_{est}$ is then given by:

$$D_{est} = P_q \cdot R_{est}$$

*Method 1*: This method finds the reference part whose pixel-based measurement is closest to the query part's $P_q$. The algorithm iterates through the reference parts and calculates the difference between $P_q$ and the reference parts' pixel values. It selects the nearest reference part and uses its conversion factor ($R_{est}$):

$$j = arg_i \, min \, |P_q - P_i|$$

$$R_{est} = R_j$$

*Method 2*: In this method, the reference pixel values ($P_i$ and $P_{i+1}$) closest to the pixel value of the query part ($P_q$) are identified. Then, the corresponding conversion factors ($R_i$ and $R_{i+1}$) are used to estimate the $R_{est}$ value through interpolation. The interpolation formula is:

$$R_{est} = \frac{R_i \cdot (P_{i+1} - P_q) + R_{i+1} \cdot (P_q - P_i)}{(P_{i+1} - P_q)}$$

Using the interpolated $R_{est}$, the estimated diameter is given by:

$$D_{est} = R_{est} \cdot P_q$$

The flowchart of the proposed R-based method is presented in Figure 4.

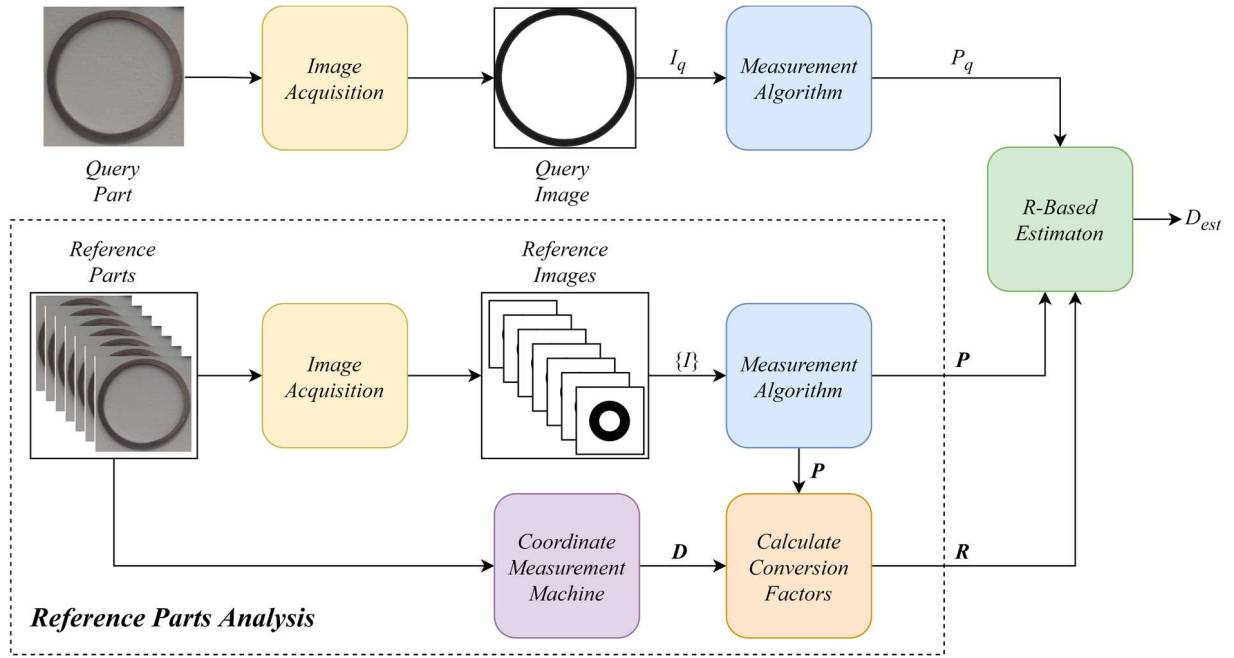

**Figure 4.** R-Based approach flowchart

### 4.2.2 P-Based Estimation Methods:

In the second approach, the diameter ($D_{est}$) is directly predicted without estimating the $R$ value, simplifying the process. Similar to the first approach, four reference gauges are used to estimate unknown diameters based on pixel-based measurements ($P$). Below are two proposed methods for prediction:

*Method 3*: This method directly estimates the diameter ($D_{est}$). The reference parts ($P_i$ and $P_{i+1}$) closest to the pixel size of the query image ($P_q$) are identified. Then, the known diameters ($D_i$ and $D_{i+1}$) of these references are used to interpolate and estimate the actual size ($D_{est}$) of the query part. The interpolation formula is:

$$D_{est} = \frac{D_i \cdot (P_{i+1} - P_q) + D_{i+1} \cdot (P_q - P_i)}{(P_{i+1} - P_q)}$$

*Method 4*: In this method, a linear relationship is established between the pixel-based measurements ($P$) and the known diameters ($D$) of the reference parts using the least-squares method. The linear equation is expressed as

$$D = m \cdot P + b$$

where $m$ represents the slope, and $b$ is the intercept of the linear equation. To determine these coefficients, the mean values of both pixel-based measurements ($P$) and known diameters ($D$) are calculated. The mean pixel value ($\overline{P}$) is computed as the average of all pixel measurements, while the mean diameter ($\overline{D}$) is obtained by averaging all diameter values.

Next, the numerator and denominator required for the slope calculation are initialized to zero. For each reference part, the difference between the pixel value and its mean ($P_i - \overline{P}$) is multiplied by the difference between the diameter value and its mean ($D_i - \overline{D}$), and the result is added to the numerator. Simultaneously, the square of the difference between the pixel value and its mean is added to the denominator. Once these summations are complete, the slope ($m$) is calculated as the ratio of the numerator to the denominator:

$$m = \frac{\sum_{i=1}^{n} (P_i - \overline{P}) \cdot (D_i - \overline{D})}{\sum_{i=1}^{n} (P_i - \overline{P})^2}$$

The intercept ($b$) is then calculated by substituting the mean values into the equation:

$$b = \overline{D} - m \cdot \overline{P}$$

Using the determined coefficients $m$ and $b$, the estimated diameter ($D_{est}$) for the query part can be calculated based on its pixel-based measurement ($P_q$):

$$D_{est} = m \cdot P_q + b$$

For clarity, the proposed P-based approach is presented in Figure 5.

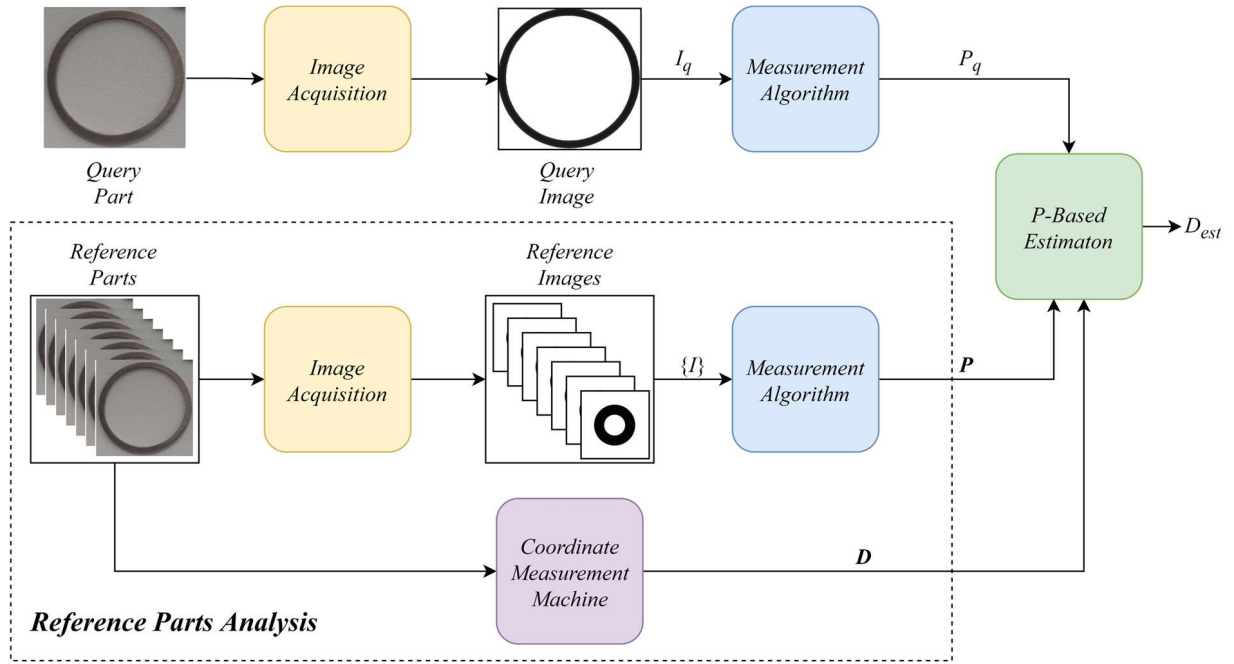

**Figure 5.** P-Based approach flowchart

## 5. Experimental Setup
### 5.1 Configuration

In this study, an experimental setup designed for industrial image processing applications was utilized (Figure 6). The setup consisted of an industrial-grade camera (Matrix Vision, Sensor IMX256, 5MP, Global Shutter, Grayscale), a telecentric illumination system, and a telecentric lens. Telecentric lenses are particularly preferred in precision measurement applications as they minimize fisheye distortion, enabling sharp and accurate imaging. These lenses are especially effective in capturing the boundary regions of parts more clearly, thereby improving measurement accuracy. To accommodate workpieces or reference gauges, a glass plate was placed between the camera and the illumination. This plate facilitated the transmission of light from the illumination to the camera through the workpiece. In setups where the illumination and the camera are positioned opposite each other, shadowing techniques can be easily implemented. Using such techniques, the setup produced images with a black part against a white background, simplifying the identification of part boundaries. Telecentric lenses operate at a fixed working distance, and deviations from this distance result in focus blur. In this study, the working distance and camera alignment were manually adjusted to ensure sharp imaging. Researchers seeking an automated focusing solution can refer to [56].

A sample image captured with the experimental setup is shown in Figure 6. Using the shadowing technique, the obtained image features a white background with a black part, which simplifies the identification of the part region. All tests were conducted using this specific setup. To minimize undesirable noise such as dust and oil, the equipment was thoroughly cleaned before each measurement, ensuring the images were suitable for accurate analysis. To eliminate potential mechanical deviations, the perpendicular alignment of the glass, camera, and illumination was carefully adjusted. Additionally, the focus was manually fine-tuned to achieve the sharpest possible boundary of the part.

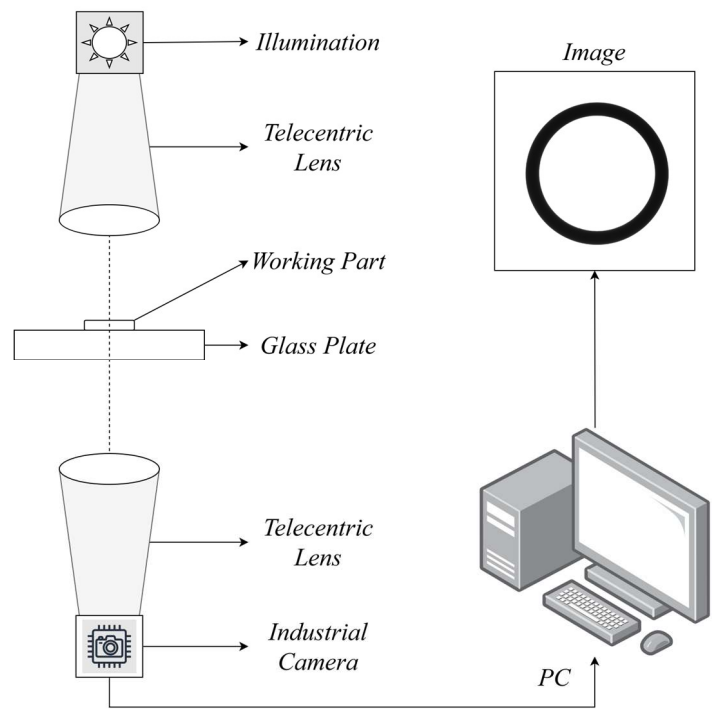

**Figure 6.** Image acquisition system.

### 5.2 Preprocessing

In industrial applications, undesired contaminants such as dirt, oil, and dust from the environment often appear in images, negatively impacting measurement accuracy. Despite careful cleaning, even a small residue can appear as a black spot in the image, leading to erroneous measurements during edge detection or pixel counting steps. To eliminate these unwanted artifacts, a connected component-based preprocessing step was applied to each image. The largest object in the image is assumed to be the workpiece under measurement. Subsequently, all connected components in the image are identified, and a mask image is created to retain the positions of the largest object and other components. The values of smaller components (noise) are then set to zero. Finally, the original image is masked with the generated mask image, ensuring that only the workpiece remains while noise is eliminated. This process effectively reduces measurement errors caused by unwanted noise in the images.

*5.3  Data Set*
*5.3.1   Synthetic Images*

In this study, initial tests were performed on synthetic images, where diameters were expressed in pixel units. Since these images lacked real-world millimeter values, the proposed methods could not be applied directly. Instead, *R* values were calculated by assuming proportional diameter sizes, allowing the behavior of the *R*-Diameter curve to be demonstrated. Synthetic images were generated following the approach in [7], with circle diameters of 50, 75, 100, 150, 200, 250, 300, 400, and 600 pixels. To simulate real-world defocus blur, a Gaussian blur filter with a σ value of 1.5 was applied to these images. Each circle image was measured using the specified algorithms, yielding diameter measurements in pixel units. These measurements were then scaled using the hypothetical sizes to calculate the corresponding *R* values.

*5.3.2   Glass References*

To evaluate the effectiveness of the proposed approach, certified reference glass with specially manufactured features was used. This reference glass contains nine circles with known diameters ranging from 1 to 12 mm (Figure 7), with an uncertainty level of 0.1 μm for each circle. During the image acquisition process, each circle was centered, and images were captured using the experimental setup (Figure 7). Undesirable noise and irrelevant circles were removed from the images through connected component analysis (Preprocessing step).

For each image capture process, the reference glass was repositioned in the setup to ensure consistent measurements. To minimize external noise and improve repeatability, ten images were captured for each diameter. The diameters were estimated in pixel units, and the average value was calculated to account for mechanical variability. In total, 90 images were obtained, which were divided into two datasets for training and testing. Specifically, diameters of 1, 3, 6, and 12 mm were used for training the proposed models, while diameters of 1.5, 2, 4, 5, and 8 mm were used for testing.

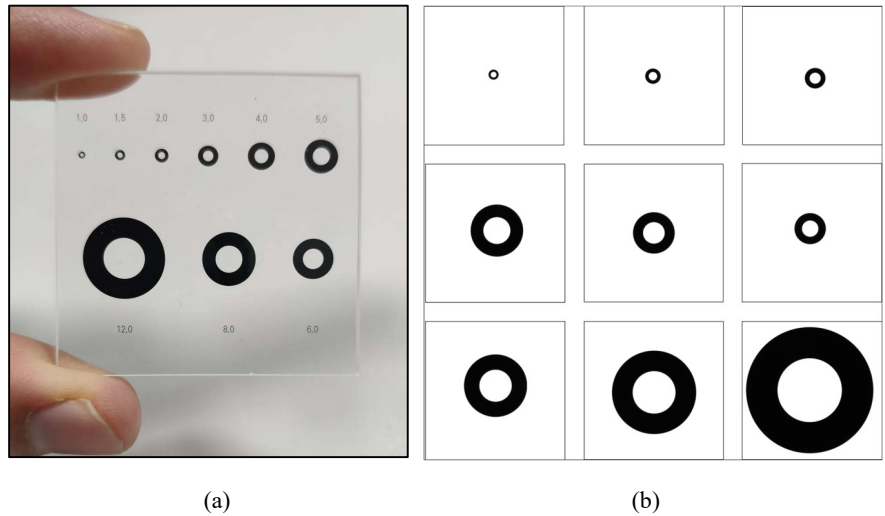

(a)            (b)

**Figure 7.** Reference glass and acquired images. (a) Glass references  (b) Digital images of glass references

*5.3.3   Real Parts*

To enable comprehensive testing of the algorithms, a dataset consisting of real workpieces was created (Figure 8). The dataset included nine disk-shaped workpieces of varying diameters. Each workpiece was measured using a Coordinate Measurement Machine (CMM) to determine its actual dimensions (Table 6). The measurement uncertainty of the CMM device is 1 μm. However, variations in the geometry of the workpieces can cause additional deviations. For instance, conical-shaped workpieces may lead to deviations of approximately ±5 μm. Despite this, these deviations were considered negligible given the substantial improvements achieved by the proposed algorithm. To eliminate potential thickness-related deviations, all workpieces underwent machining, reducing their thickness to 0.9 ± 0.01 mm. Similar to the glass reference test, the workpieces were divided into two sets for training and testing. Four workpieces were allocated to the training set, while five were reserved for testing. During image acquisition, each workpiece's image was captured ten times to enhance repeatability and minimize variability. Before each capture, the workpiece was removed from the setup and repositioned. The average of the obtained measurements was calculated to reduce repeatability errors. In total, 90 images were captured across all workpieces.

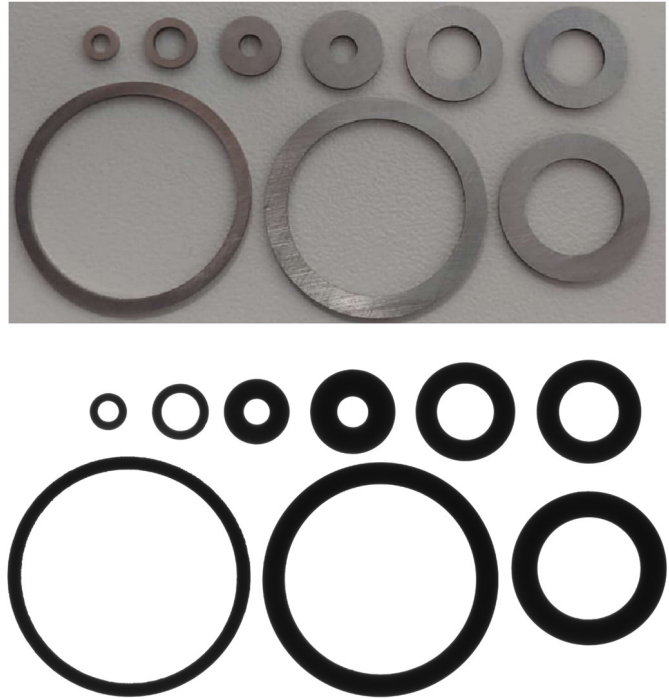

**Figure 8.** Metal parts and their acquired images

*5.4 Measurement Algorithms*

In camera-based measurement systems, particularly in applications requiring high precision for parts with tight tolerances, sub-pixel-based methods are often preferred. This study focuses on diameter measurement and utilizes the most effective diameter measurement algorithms from the literature in the conducted tests. Sub-pixel-based measurement algorithms typically follow two main approaches: edge-based and non-edge-based methods. The edge-based approach involves detecting edge points in the image using sub-pixel algorithms and then calculating the diameter by applying least-square fitting to the detected points. In this study, two of the best edge detection methods from the literature Devernay [8] and Sub-pixel Edge [7] have been selected as the first two algorithms. The third algorithm chosen is the Sub-pixel Counting method [33], which is known for its speed advantage, particularly in diameter measurement. Unlike edge-based methods, the [33] method focuses on area detection rather than edge detection. Area detection is performed by summing the black and gray regions in the image with certain weighting ratios, followed by diameter estimation.

## 6. Experimentals Results

In this study, two approaches, termed *R*-based and *P*-based, were proposed to address the problem of diameter estimation. These approaches are designed to integrate seamlessly with existing diameter measurement algorithms from the literature. Each approach includes two sub-methods. To evaluate the effectiveness of these proposed approaches and sub-methods, experiments were conducted on two datasets: one comprising glass samples and the other containing metal samples. Both datasets consist of nine samples, with four designated for training and five for testing. Each dataset was evaluated using three distinct diameter measurement algorithms.

*6.1 Synthetic Images*

To assess the performance of the best-performing diameter measurement algorithms for small diameters, a basic simulation study was conducted. Synthetic circles with diameters ranging from 50 to 600 pixels were generated based on the methodology of [7]. To simulate a focus effect, a Gaussian blur filter with a ratio of 1.5 was applied to the images. The diameters of these circles were then calculated in pixels using algorithms from [7], [8], and [33], with the percentage errors summarized in Table 2.

**Table 2:** Percentage error for simulations (synthetic images)

| Diameters (px) | 50 | 75 | 100 | 200 | 300 | 400 | 600 |
|---|---|---|---|---|---|---|---|
| Devernay [8] | 0.294 | 0.136 | 0.073 | 0.018 | 0.008 | 0.004 | 0.002 |
| Subpixel Edges [7] | 0.176 | 0.130 | 0.028 | 0.011 | 0.001 | 0.000 | 0.002 |
| Subpixel Counting [33] | 0.345 | 0.010 | 0.008 | 0.005 | 0.009 | 0.031 | 0.001 |

When examining Table 2, it is evident that the methods become increasingly unstable as the diameter size decreases. This behavior is consistent with previously reported findings in the literature (Figure 9). Such instability highlights the need for advanced systems and methods to correct these errors. In many applications, a percentage error of around 1% may be acceptable. However, in precision measurement applications, stricter tolerances are required, especially for small workpieces where narrower tolerances make accurate measurement critical. Therefore, the development of robust methods to minimize these errors is essential.

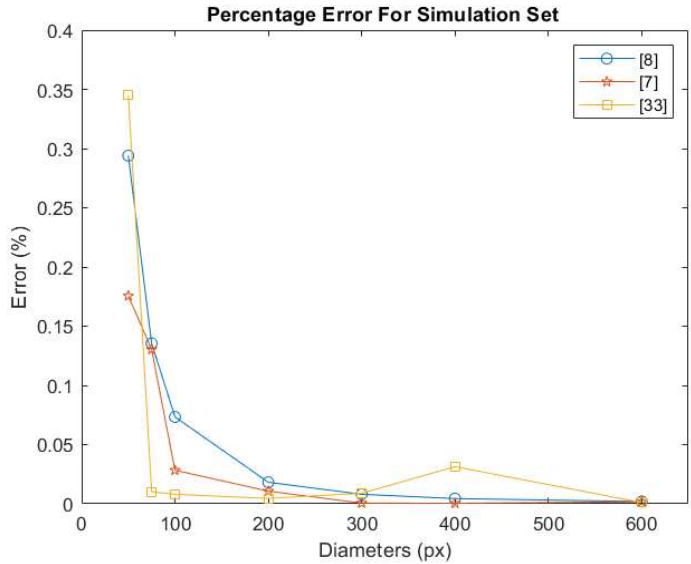

**Figure 9.** Percentage error for synthetic images (disk)

*6.2 Glass References*

The first dataset consisted of specially manufactured glass references (Figure 3). For each circle on the glass, ten images were captured, and their diameters were calculated in pixel units using algorithms [7], [8], and [33]. To minimize repeatability errors, the average of the ten measurements was calculated for each reference. Using these average values and the known true diameters $D$, $R$ values were calculated. The $R$, $D$, and $P$ values obtained from the reference diameters (1, 3, 6, 12 mm) were used to conduct tests for the test diameters (1.5, 2, 4, 5, 8 mm). For each approach, the difference between the predicted and actual measurements was calculated, followed by determining the percentage error. This process was repeated for all three measurement algorithms, and the results were summarized in Tables 3, 4, and 5. The mean absolute percentage errors (MAPE) for the glass dataset were then compared (Figure 10).

**Table 3**. Mean absolute percentage error values obtained for glass samples using the [8] method.

| | Methods | Absolute Percentage Error (%) | | | | |
| --- | --- | --- | --- | --- | --- | --- |
| | | R-Based Approach | | | P-Based Approach | |
| | | M0 | M1 | M2 | M3 | M4 |
| Diameters [mm] | 1.5 | 1.1687 | 1.6794 | **0.8755** | 0.0469 | **0.0274** |
| | 2 | **0.3725** | 0.7384 | 0.8784 | **0.0507** | 0.0619 |
| | 4 | 0.7172 | 0.3473 | **0.1119** | **0.0052** | 0.0109 |
| | 5 | 0.9310 | 0.1476 | **0.0884** | 0.0049 | **0.0034** |
| | 8 | 1.2766 | 0.1943 | **0.0670** | 0.0040 | **0.0003** |
| | MAPE | 0.8932 | 0.6214 | **0.4042** | 0.0223 | **0.0208** |

**Table 4.** Mean absolute percentage error values obtained for glass samples using the [7] method.

| | | Absolute Percentage Error (%) | | | | |
| --- | --- | --- | --- | --- | --- | --- |
| | | R-Based Approach | | | P-Based Approach | |
| | Methods | M0 | M1 | M2 | M3 | M4 |
| Diameters [mm] | 1.5 | 1.3079 | 1.8274 | **0.4464** | 0.0363 | **0.0308** |
| | 2 | **0.4378** | 0.8244 | 0.4442 | **0.0392** | 0.0517 |
| | 4 | 0.7806 | 0.3893 | **0.1227** | 0.0118 | **0.0020** |
| | 5 | 1.0270 | 0.1673 | **0.1001** | **0.0075** | 0.0101 |
| | 8 | 1.4169 | 0.2180 | **0.0746** | **0.0002** | 0.0023 |
| | MAPE | 0.9940 | 0.6853 | **0.2376** | **0.0190** | 0.0194 |

**Table 5.** Mean absolute percentage error values obtained for glass samples using the [33] method.

| | | Absolute Percentage Error (%) | | | | |
| --- | --- | --- | --- | --- | --- | --- |
| | | R-Based Approach | | | P-Based Approach | |
| | Methods | M0 | M1 | M2 | M3 | M4 |
| Diameters [mm] | 1.5 | 0.8726 | 1.2341 | **0.7277** | 0.0948 | **0.0807** |
| | 2 | **0.2773** | 0.5484 | 0.7104 | **0.0787** | 0.0914 |
| | 4 | 0.5276 | 0.2543 | **0.0963** | **0.0090** | 0.0208 |
| | 5 | 0.6799 | 0.1033 | **0.0729** | 0.0032 | **0.0018** |
| | 8 | 0.9409 | 0.1557 | **0.0555** | 0.0066 | **0.0037** |
| | MAPE | 0.6597 | 0.4592 | **0.3326** | **0.0385** | 0.0397 |

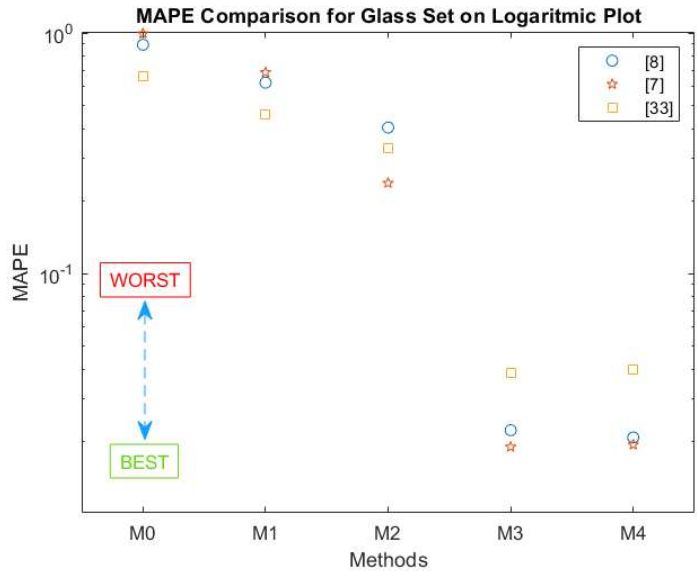

**Figure 10.** Mean absolute percentage errors according to proposed approaches and estimation algorithms applied on "Glass" set

When examining Tables 3, 4, and 5, it is evident that the diameter measurement methods yield similar trends, though with slight differences in precision. Across all methods, the calculated $R$ values decrease as the diameter increases, ultimately converging towards a specific point. The observations from the glass dataset indicate that all proposed sub-methods outperform the base method in terms of percentage error. Among these, the $P$-based approaches exhibit overall better performance compared to the $R$-based approaches. Within the $P$-based methods, the results are highly consistent, showing negligible differences between sub-methods. Notably, when paired with estimation algorithm [7], Method 3 achieves the best performance for the glass dataset. Additionally, the $R$ values are observed to decrease as the diameter increases, which is consistent across all methods. This trend reflects the convergence of $R$ towards a specific value for larger diameters, as illustrated in Figure 2. These findings highlight the robustness and reliability of the proposed $P$-based approaches for precise diameter estimation.

### 6.3 Real Parts

The second dataset consisted of nine different diameter metal washers, used as metal workpieces. The actual measurements $D$ (in millimeters) for each washer were obtained using a CMM. Subsequently, ten images were captured for each part using the experimental setup. The diameters in the captured images were calculated in pixel units using algorithms [7], [8], and [33]. To minimize repeatability errors, the average of the ten measurements was computed for each reference. Using these average values and the known true diameters, the $R$ values were calculated. The $R$ values obtained from the reference diameters (3.665, 5.800, 8.594, and 24.320 mm) were utilized to apply the proposed approach to the test set, which included diameters of 5.398, 6.431, 10.458, 11.046, and 23.668 mm. The difference between the predicted and actual measurements was calculated, followed by determining the percentage error. This process was repeated for all three measurement algorithms, and the results are summarized in Tables 6, 7, and 8. The mean absolute percentage errors (MAPE) for the metal dataset were compared (Figure 11).

**Table 6.** Mean absolute percentage error values obtained for metal parts using the [8] method.

| | | Absolute Percentage Error (%) | | | | |
|---|---|---|---|---|---|---|
| | | R-Based Approach | | | P-Based Approach | |
| | Methods | M0 | M1 | M2 | M3 | M4 |
| Diameters [mm] | 5.398 | 0.4391 | 0.3995 | **0.1053** | 0.0718 | **0.0296** |
| | 6.431 | 0.3529 | 0.3529 | **0.0169** | 0.1440 | **0.1201** |
| | 10.458 | 0.1316 | 0.1316 | **0.0634** | 0.0310 | **0.0066** |
| | 11.046 | 0.1472 | 0.1159 | **0.0576** | 0.0214 | **0.0080** |
| | 23.668 | 0.4147 | 0.1509 | **0.0387** | 0.0433 | **0.0280** |
| | MAPE | 0.2971 | 0.2302 | **0.0564** | 0.0623 | **0.0385** |

**Table 7.** Mean absolute percentage error values obtained for metal parts using the [7] method.

| | | Absolute Percentage Error (%) | | | | |
|---|---|---|---|---|---|---|
| | | R-Based Approach | | | P-Based Approach | |
| | Methods | M0 | M1 | M2 | M3 | M4 |
| Diameters [mm] | 5.398 | 0.5008 | 0.4842 | **0.1386** | 0.0692 | **0.0268** |
| | 6.431 | 0.3886 | 0.3886 | **0.0459** | 0.1429 | **0.1189** |
| | 10.458 | 0.1495 | 0.1495 | **0.0686** | 0.0250 | **0.0058** |
| | 11.046 | 0.1731 | 0.1392 | **0.0667** | 0.0186 | **0.0056** |
| | 23.668 | 0.4827 | 0.1695 | **0.0390** | 0.0436 | **0.0284** |
| | MAPE | 0.3389 | 0.2662 | **0.0718** | 0.0599 | **0.0371** |

**Table 8.** Mean absolute percentage error values obtained for metal parts using the [33] method.

| | | Absolute Percentage Error (%) | | | | |
|---|---|---|---|---|---|---|
| | | R-Based Approach | | | P-Based Approach | |
| | Methods | M0 | M1 | M2 | M3 | M4 |
| Diameters [mm] | 5.398 | 0.3367 | 0.2410 | **0.0384** | 0.0851 | **0.0356** |
| | 6.431 | 0.3155 | 0.3155 | **0.0606** | 0.1731 | **0.1480** |
| | 10.458 | 0.1080 | 0.1080 | **0.0572** | 0.0300 | **0.0045** |
| | 11.046 | 0.1099 | 0.0859 | **0.0432** | **0.0132** | 0.0165 |
| | 23.668 | 0.3327 | 0.1364 | **0.0263** | 0.0317 | **0.0115** |
| | MAPE | 0.2405 | 0.1774 | **0.0452** | 0.0666 | **0.0432** |

When examining Tables 6, 7, and 8, it can be observed that the $R$ values decrease as the diameter increases, similar to the trend in the glass dataset. The convergence of $R$ values towards a specific point is evident in both datasets. The observations from the metal dataset reveal that the $P$-based approach outperforms the $R$-based approach, consistent with the findings from the glass dataset. However, unlike the glass dataset, notable differences are observed among methods 3 and 4 within the $P$-based approach. Among these, Method 4 demonstrates the best performance across all estimation algorithms. Furthermore, the combination of the [7] algorithm with Method 4 achieves the lowest error for the metal dataset, highlighting its effectiveness in diameter estimation.

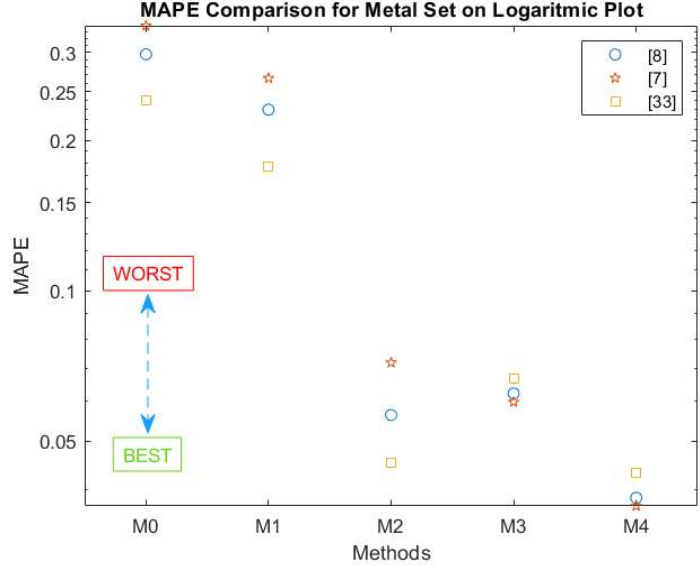

**Figure 11.** Mean absolute percentage errors according to proposed approaches and estimation algorithms applied on "Metal" set

An analysis of the error reveals that the metal dataset contains samples ranging from 3 to 24 mm, while the glass dataset includes smaller samples ranging from 1 to 12 mm. Smaller diameters tend to produce higher errors, which significantly affect the overall average. Consequently, the mean absolute percentage error values for the metal parts are lower compared to the glass parts, as the larger diameters in the metal dataset contribute to more stable and accurate measurements. Overall, the tests conducted on both datasets demonstrate the effectiveness of the proposed approaches. The results confirm that the $P$-based methods, particularly Method 4, provide superior performance for diameter estimation in both metal and glass datasets.

*6.4 Comparison*

The performance of the proposed methods was evaluated across both the glass and metal datasets. Among the methods, Method 4 consistently demonstrated the lowest error percentage for the metal dataset (Figure 11). Similarly, in the glass dataset, the $P$-based methods significantly outperformed the $R$-based methods, with all $P$-based methods showing comparable results (Figure 10). Consequently, Method 4 was selected as the best-performing approach.

To further validate its performance, Method 4 was tested in conjunction with the diameter estimation algorithms [7], [8], and [33] and compared to the base method (Table 9). The evaluations included both absolute percentage error and absolute error values for the glass and metal datasets. Across all tests, the proposed approach effectively eliminated the error caused by the base method, regardless of the estimation algorithm used (Figures 14a, 14b). Furthermore, it is demonstrated in Figures 12 and 13 that the proposed method improves each measurement algorithm individually.

**Table 9.** Absolute errors for proposed approach. (The '+' sign was used to indicate that the proposed approach is employed in conjunction with the diameter measurement algorithm.)

| | Absolute Error (μm) | | | | | | | | | |
|---|---|---|---|---|---|---|---|---|---|---|
| Method | Test Diameters for Metal Samples [μm] | | | | | Test Diameters for Glass Samples [μm] | | | | |
| | 5400 | 6430 | 10460 | 11050 | 23670 | 1500 | 2000 | 4000 | 6000 | 8000 |
| [8] + M0 | 23.70 | 22.70 | 13.76 | 16.26 | 98.16 | 17.53 | 7.45 | 28.69 | 46.55 | 102.13 |
| [8] + M4 | 1.60 | 7.73 | 0.69 | 0.88 | 6.64 | 0.41 | 1.24 | 0.44 | 0.17 | 0.02 |
| [7] + M0 | 27.03 | 24.99 | 15.63 | 19.12 | 114.25 | 19.62 | 8.76 | 31.23 | 51.35 | 113.35 |
| [7] + M4 | 1.45 | 7.65 | 0.60 | 0.62 | 6.71 | 0.46 | 1.03 | 0.08 | 0.51 | 0.19 |
| [33] + M0 | 18.18 | 20.29 | 11.29 | 12.14 | 78.74 | 13.09 | 5.55 | 21.10 | 34.00 | 75.27 |
| [33] + M4 | 1.92 | 9.52 | 0.47 | 1.82 | 2.73 | 1.21 | 1.83 | 0.83 | 0.09 | 0.30 |

In the test diameters, the base method produced errors ranging from 8 μm to 113 μm, while the proposed approach significantly reduced these errors to the range of 1–2 μm. This improvement highlights the robustness of Method 4 in achieving precise measurements. The glass dataset, with its specially manufactured samples and tight production tolerances of approximately 0.1 μm, provided controlled conditions that allowed for clearer and more consistent results. These conditions enabled the proposed approach to excel in accurately estimating diameters, demonstrating its remarkable success in reducing measurement errors across both the glass and metal datasets.

In contrast, the metal dataset presented additional challenges due to inherent variability in the workpieces. The samples in the metal dataset, while having the same nominal thickness, exhibited differences in their physical characteristics, which introduced variability. Furthermore, the known dimensions of the metal parts were determined using a CMM device, whose measurement uncertainty also influenced the results. Despite these challenges, the proposed approach performed effectively, achieving a maximum error of approximately 6 μm, which is a significant improvement over the base method.

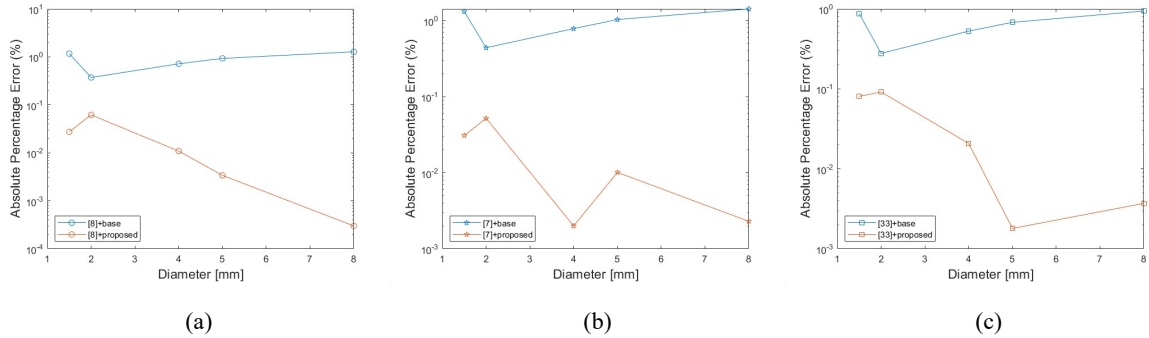

**Figure 12.** Performance indicators of the proposed algorithm on the glass dataset in terms of MAPE (Mean Absolute Percentage Error) for each measurement algorithm. (a) represents the results for method [8], (b) for method [7], and (c) for method [33].

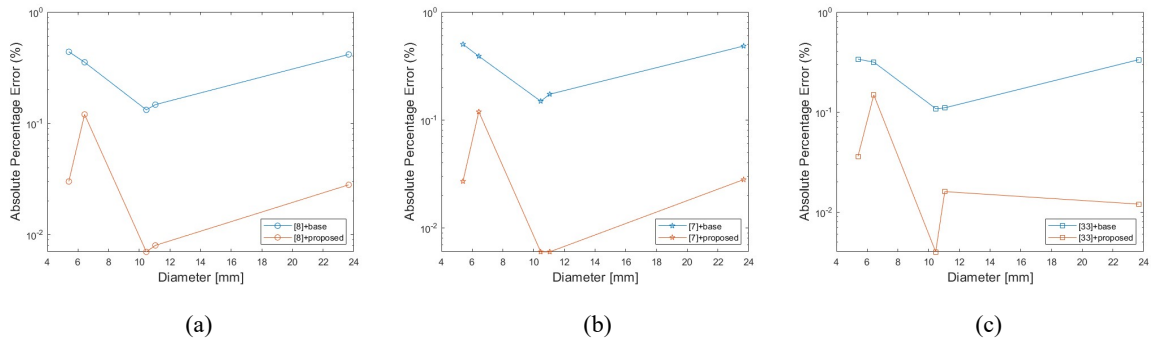

**Figure 13.** Performance indicators of the proposed algorithm on the metal dataset in terms of MAPE (Mean Absolute Percentage Error) for each measurement algorithm. (a) represents the results for method [8], (b) for method [7], and (c) for method [33].

When evaluating the diameter estimation algorithms, all three algorithms ([7], [8], and [33]) demonstrated similar trends in performance. The combination of the [7] algorithm with Method 4 consistently yielded the best results across both datasets, further validating the robustness and adaptability of the proposed approach. The comparison between the glass and metal datasets also highlights the impact of sample characteristics on measurement precision. The glass dataset, with smaller diameters (1–12 mm), was more susceptible to errors, while the metal dataset, with larger diameters (3–24 mm), benefited from increased stability and lower error rates (Figure 14). Overall, the proposed approach effectively mitigates the limitations of the base method and exhibits exceptional performance, particularly when combined with well-suited estimation algorithms. These results underscore the reliability of the proposed methods for precise diameter estimation across a variety of datasets and conditions.

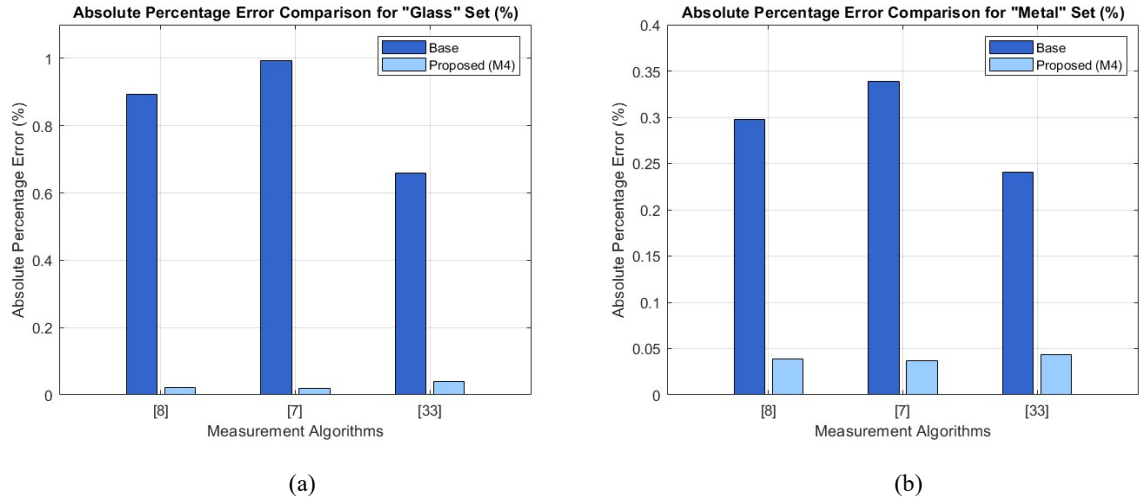

**Figure 14.** Absolute Percentage Error of the proposed method across measurement algorithms. (a) Results for the glass dataset, (b) Results for the metal dataset

7. **Discussion**

In industrial camera measurement applications, ensuring workpieces meet tolerance limits is critical. As tolerances narrow, the equipment and algorithms used in measurements may require refinement. For instance, telecentric lenses are often employed to achieve sharper image transitions, and sub-pixel-based measurement algorithms are preferred to enhance measurement accuracy. However, despite these advancements, measurement errors can still arise due to limitations in the measurement setup and algorithms. Tests conducted with two datasets and three diameter measurement algorithms reveal that smaller diameters tend to produce higher $R$ values (Figure 2). This behavior is rarely encountered in standard applications, as these typically involve measuring a single diameter size with a custom reference gauge tailored to that specific workpiece. However, when measuring workpieces of varying diameters using the same setup, discrepancies arise. If a sufficient field of view exists, a new reference gauge is typically produced for the new workpiece dimensions. Otherwise, the existing $R$ value is used, leading to the measurement error shown in Figure 1.

To address this issue, this study proposes a novel approach that significantly reduces measurement errors. The proposed method enables accurate measurement of multiple diameters within the field of view using only a few reference parts. This approach offers originality and innovation, as it eliminates errors inherent in existing systems and is directly applicable to industrial settings. A detailed analysis was performed using two datasets Glass and Metal, and two primary approaches: $R$-based and $P$-based. Each approach included two sub-methods, all tested on the specified datasets. Results indicate that the $P$-based approach outperforms the $R$-based approach. The behavior of the $P$-based approach allows for more accurate modeling compared to the $R$-based approach, where curves tend to converge toward a single point.

In comparing the datasets, measurements in the Glass dataset yielded better results than the Metal dataset. The Glass dataset was specifically produced, minimizing errors originating from the samples, while the Metal dataset included workpieces with inherent variations such as conicity. Additionally, real dimensions of metal parts were measured using a CMM device, introducing measurement uncertainty. These factors underscore the advantage of using specially produced reference gauges, as seen in the Glass dataset.

Figure 2 illustrates that while the Glass and Metal datasets exhibit similar behavior, the Glass dataset consistently produces lower values at equivalent diameters. This difference arises from the thickness variations between the metal and glass samples, with thicker metal parts affecting image focus. Future studies could further investigate the impact of thickness differences on diameter measurements to refine the proposed approach.

The proposed method demonstrated significant error reductions in both datasets. In the Glass dataset, errors decreased from 13–113 μm to approximately 1 μm, while in the Metal dataset, errors reduced from 18–114 μm to a maximum of 9 μm. This improvement was consistent across the three tested diameter measurement algorithms, validating the method's success.

The study employed four training samples, with Method 4 emerging as the most successful. This method requires at least two references, allowing researchers to implement the approach with only two reference gauges if desired. The use of specially produced gauges further enhances the method's accuracy. Additionally, Method 4 enables estimation of diameters beyond the known references, increasing its practicality.

While the requirement for at least two reference gauges may be seen as a limitation, the proposed method offers several advantages. It eliminates measurement errors inherent in existing systems, enables measurement of parts with varying diameters without mechanical system modifications, and reduces costs associated with producing reference gauges. For small diameter parts where special gauges may be impractical, the method allows measurements using larger reference diameters.

The innovation of this study lies in its ability to eliminate diameter measurement errors for circular parts, applicable to any measurement algorithm. Testing with both synthetic and real datasets and detailed analysis of mm/px variance according to diameters further highlight the novelty of this work. By leveraging a few known reference values, the method enables high-precision measurements for all parts within the camera's field of view, reducing reliance on costly reference parts while achieving exceptional accuracy.

8. Conclusion

This study introduces a novel approach for correcting diameter measurement errors in industrial applications, particularly addressing the challenges of measuring parts with narrow tolerances that demand high precision. The proposed approach enhances the performance of existing diameter measurement algorithms by utilizing the most advanced methods available in the literature. To facilitate precise measurements in industrial settings, the experimental setup was optimized for shadowing techniques with telecentric lenses, and mechanical components, such as lens alignment, were meticulously adjusted to minimize potential errors. Despite these efforts, measurement errors stemming from both mechanical and software-related factors persist in precision measurement applications. These errors are influenced by the measurement diameter. In this study, the primary focus was to identify and mitigate measurement errors caused by the camera system, encompassing both hardware and software aspects. The experimental setup included telecentric lenses, an industrial-grade camera, and datasets comprising a specially manufactured glass reference surface and metal workpieces. The glass dataset featured samples with diameters ranging from 1 to 12 mm and near-perfect roundness, while the metal dataset included workpieces produced through mass manufacturing, with diameters ranging from 3 to 24 mm. Two approaches, $R$-based and $P$-based, were proposed and tested on both datasets. Among nine samples in each dataset, four were used for training and five for testing. The $R$-based approach estimated unknown diameter values using known $R$ values, while the $P$-based approach directly estimated unknown diameters in millimeters using known pixel-based diameter values. Results indicate that the $P$-based approach significantly outperforms the $R$-based approach, reducing measurement errors from 3–114 μm to 1–2 μm. When combined with the proposed method, all tested diameter measurement algorithms exhibited similar levels of improvement, demonstrating the versatility of the proposed approach. Notably, the method outlined in [7] achieved the best results when paired with this approach. This study provides an effective solution for mitigating errors in the measurement of tight-tolerance parts. Furthermore, it enables accurate measurements of workpieces with varying diameters using a single experimental setup while maintaining low error rates. The findings emphasize the potential for integrating the proposed method into industrial applications to enhance measurement accuracy, reduce costs, and streamline the measurement process.

**Data Availability**
Data will be made available on reasonable request.

**Funding**
This work was supported by the Scientific And Technological Research Council Of Türkiye [Project No. 123E343].

**Contributions**
All the Authors have contributed equally.

**Conflict of interest**
The author declares that he has no conflicts of interest to disclose.

**Acknowledgments**
We thank the anonymous referees for their useful suggestions

**APPENDIX**

This appendix provides detailed tables summarizing the pixel-based measurement results, averages, and calculated $R$ values for both the glass and metal datasets. The tables are divided into two groups based on the datasets: **A1, A2, A3** present the results for the glass dataset, while **B1, B2, B3** correspond to the results for the metal dataset. Each group includes the results obtained using three different measurement algorithms ([7], [8], and [33]). The tables list the measurements for all reference and test parts in pixel units, along with their averages and the calculated $R$ values. These values were subsequently used in the proposed approaches for diameter estimation.

|  | | Diameter Measurements [px] | | | | | | | | |
|---|---|---|---|---|---|---|---|---|---|---|
|  | Actual Values | 1 mm | 1.5 mm | 2 mm | 3 mm | 4 mm | 5 mm | 6 mm | 8 mm | 12 mm |
| Measurements | 1 | 49.39 | 75.32 | 101.21 | 152.91 | 204.65 | 256.36 | 308.05 | 411.64 | 618.62 |
|  | 2 | 49.37 | 75.34 | 101.23 | 152.89 | 204.66 | 256.36 | 308.11 | 411.62 | 618.56 |
|  | 3 | 49.31 | 75.31 | 101.25 | 152.96 | 204.69 | 256.39 | 308.13 | 411.65 | 618.59 |
|  | 4 | 49.40 | 75.32 | 101.22 | 152.97 | 204.66 | 256.39 | 308.13 | 411.59 | 618.56 |
|  | 5 | 49.40 | 75.29 | 101.25 | 152.95 | 204.66 | 256.38 | 308.11 | 411.58 | 618.58 |
|  | 6 | 49.41 | 75.32 | 101.24 | 153.00 | 204.70 | 256.38 | 308.10 | 411.60 | 618.57 |
|  | 7 | 49.39 | 75.29 | 101.20 | 152.99 | 204.69 | 256.35 | 308.08 | 411.65 | 618.56 |
|  | 8 | 49.38 | 75.31 | 101.23 | 153.00 | 204.67 | 256.40 | 308.14 | 411.59 | 618.59 |
|  | 9 | 49.38 | 75.33 | 101.19 | 153.01 | 204.60 | 256.38 | 308.12 | 411.59 | 618.58 |
|  | 10 | 49.35 | 75.30 | 101.24 | 153.01 | 204.67 | 256.38 | 308.09 | 411.56 | 618.53 |
|  | Avg(px) | 49.38 | 75.31 | 101.23 | 152.97 | 204.67 | 256.38 | 308.11 | 411.61 | 618.57 |
|  | R | 20.251 | 19.917 | 19.758 | 19.612 | 19.544 | 19.503 | 19.474 | 19.436 | 19.400 |

**Table A1**. Pixel-based diameter measurement results obtained for the glass set using the [8] method.

|  | | Diameter Measurements [px] | | | | | | | | |
|---|---|---|---|---|---|---|---|---|---|---|
|  | Actual Values | 1 mm | 1.5 mm | 2 mm | 3 mm | 4 mm | 5 mm | 6 mm | 8 mm | 12 mm |
| Measurements | 1 | 49.14 | 75.06 | 100.94 | 152.63 | 204.36 | 256.08 | 307.78 | 411.38 | 618.36 |
|  | 2 | 49.14 | 75.09 | 100.96 | 152.62 | 204.38 | 256.07 | 307.85 | 411.36 | 618.31 |
|  | 3 | 49.08 | 75.04 | 100.97 | 152.68 | 204.39 | 256.11 | 307.86 | 411.39 | 618.34 |
|  | 4 | 49.16 | 75.07 | 100.94 | 152.69 | 204.39 | 256.11 | 307.85 | 411.32 | 618.31 |
|  | 5 | 49.16 | 75.02 | 100.98 | 152.67 | 204.38 | 256.10 | 307.84 | 411.31 | 618.32 |
|  | 6 | 49.17 | 75.05 | 100.97 | 152.73 | 204.42 | 256.10 | 307.82 | 411.33 | 618.32 |
|  | 7 | 49.14 | 75.03 | 100.93 | 152.71 | 204.41 | 256.07 | 307.80 | 411.39 | 618.31 |

| | Actual Values | 1 mm | 1.5 mm | 2 mm | 3 mm | 4 mm | 5 mm | 6 mm | 8 mm | 12 mm |
|---|---|---|---|---|---|---|---|---|---|---|
| | 8 | 49.14 | 75.05 | 100.96 | 152.72 | 204.38 | 256.12 | 307.87 | 411.32 | 618.34 |
| | 9 | 49.14 | 75.07 | 100.92 | 152.73 | 204.29 | 256.11 | 307.85 | 411.32 | 618.33 |
| | 10 | 49.10 | 75.07 | 100.97 | 152.74 | 204.38 | 256.11 | 307.82 | 411.29 | 618.28 |
| | Avg(px) | 49.14 | 75.05 | 100.95 | 152.69 | 204.38 | 256.10 | 307.83 | 411.34 | 618.32 |
| | R | 20.351 | 19.986 | 19.811 | 19.645 | 19.571 | 19.524 | 19.491 | 19.449 | 19.407 |

**Table A2**. Pixel-based diameter measurement results obtained for the glass set using the [7] method.

| | | Diameter Measurements [px] | | | | | | | | |
|---|---|---|---|---|---|---|---|---|---|---|
| | Actual Values | 1 mm | 1.5 mm | 2 mm | 3 mm | 4 mm | 5 mm | 6 mm | 8 mm | 12 mm |
| Measurements | 1 | 50.03 | 75.91 | 101.80 | 153.46 | 205.19 | 256.89 | 308.32 | 412.12 | 619.06 |
| | 2 | 49.94 | 75.90 | 101.81 | 153.36 | 205.19 | 256.88 | 308.55 | 412.07 | 618.95 |
| | 3 | 49.88 | 75.90 | 101.81 | 153.49 | 205.19 | 256.86 | 308.59 | 412.06 | 619.04 |
| | 4 | 50.00 | 75.86 | 101.74 | 153.50 | 205.16 | 256.90 | 308.60 | 412.04 | 618.96 |
| | 5 | 49.89 | 75.79 | 101.76 | 153.52 | 205.20 | 256.84 | 308.58 | 412.06 | 619.02 |
| | 6 | 50.06 | 75.87 | 101.76 | 153.54 | 205.23 | 256.87 | 308.59 | 412.08 | 618.97 |
| | 7 | 50.02 | 75.84 | 101.76 | 153.52 | 205.25 | 256.89 | 308.57 | 412.08 | 618.96 |
| | 8 | 49.92 | 75.85 | 101.78 | 153.54 | 205.19 | 256.88 | 308.63 | 412.05 | 619.02 |
| | 9 | 49.99 | 75.90 | 101.73 | 153.53 | 205.14 | 256.83 | 308.64 | 412.04 | 618.99 |
| | 10 | 49.93 | 75.90 | 101.78 | 153.52 | 205.11 | 256.87 | 308.58 | 412.01 | 618.88 |
| | Avg(px) | 49.97 | 75.87 | 101.77 | 153.50 | 205.19 | 256.87 | 308.56 | 412.06 | 618.99 |
| | R | 20.014 | 19.770 | 19.652 | 19.544 | 19.495 | 19.465 | 19.445 | 19.415 | 19.387 |

**Table A3**. Pixel-based diameter measurement results obtained for the glass set using the [33] method.

| | | Diameter Measurements [px] | | | | | | | | |
|---|---|---|---|---|---|---|---|---|---|---|
| | Actual Values | 3.665 mm | 5.398 mm | 6.431 mm | 8.594 mm | 10.458 mm | 11.046 mm | 15.8 mm | 23.668 mm | 24.32 mm |
| Measurements | 1 | 186.80 | 276.19 | 329.46 | 441.78 | 538.23 | 568.61 | 814.25 | 1221.54 | 1255.87 |
| | 2 | 186.82 | 276.14 | 329.40 | 441.71 | 538.17 | 568.56 | 814.24 | 1221.49 | 1255.84 |
| | 3 | 186.79 | 276.27 | 329.35 | 441.72 | 538.22 | 568.58 | 814.23 | 1221.53 | 1255.82 |
| | 4 | 186.76 | 276.23 | 329.33 | 441.71 | 538.11 | 568.56 | 814.17 | 1221.52 | 1255.85 |
| | 5 | 186.83 | 276.42 | 329.38 | 441.67 | 538.13 | 568.44 | 814.26 | 1221.52 | 1255.88 |
| | 6 | 186.76 | 276.18 | 329.34 | 441.76 | 538.18 | 568.64 | 814.15 | 1221.53 | 1255.85 |
| | 7 | 186.84 | 276.25 | 329.37 | 441.69 | 538.30 | 568.72 | 814.24 | 1221.52 | 1255.86 |
| | 8 | 186.80 | 276.13 | 329.38 | 441.76 | 538.33 | 568.61 | 814.17 | 1221.52 | 1255.86 |
| | 9 | 186.75 | 276.20 | 329.36 | 441.67 | 538.25 | 568.53 | 814.21 | 1221.47 | 1255.83 |
| | 10 | 186.83 | 276.25 | 329.34 | 441.62 | 538.31 | 568.46 | 814.26 | 1221.53 | 1255.88 |
| | Avg(px) | 186.80 | 276.22 | 329.37 | 441.71 | 538.22 | 568.57 | 814.22 | 1221.52 | 1255.85 |
| | R | 19.620 | 19.542 | 19.525 | 19.456 | 19.431 | 19.428 | 19.405 | 19.376 | 19.365 |

**Table B1**. Pixel-based diameter measurement results obtained for the metal set using the [8] method.

| | | Diameter Measurements [px] | | | | | | | | |
|---|---|---|---|---|---|---|---|---|---|---|
| | Actual Values | 3.665 mm | 5.398 mm | 6.431 mm | 8.594 mm | 10.458 mm | 11.046 mm | 15.8 mm | 23.668 mm | 24.32 mm |
| Measurements | 1 | 186.34 | 275.74 | 329.01 | 441.34 | 537.77 | 568.18 | 813.83 | 1221.13 | 1255.46 |
| | 2 | 186.36 | 275.69 | 328.95 | 441.25 | 537.71 | 568.13 | 813.81 | 1221.07 | 1255.43 |
| | 3 | 186.32 | 275.82 | 328.91 | 441.28 | 537.77 | 568.16 | 813.81 | 1221.13 | 1255.41 |
| | 4 | 186.31 | 275.77 | 328.88 | 441.27 | 537.66 | 568.12 | 813.75 | 1221.11 | 1255.46 |
| | 5 | 186.37 | 275.97 | 328.93 | 441.21 | 537.68 | 568.00 | 813.82 | 1221.11 | 1255.48 |

| | | | | | | | | | |
|---|---|---|---|---|---|---|---|---|---|
| | 6 | 186.29 | 275.72 | 328.87 | 441.32 | 537.72 | 568.21 | 813.73 | 1221.13 | 1255.45 |
| | 7 | 186.37 | 275.80 | 328.92 | 441.24 | 537.85 | 568.29 | 813.81 | 1221.10 | 1255.47 |
| | 8 | 186.34 | 275.68 | 328.92 | 441.31 | 537.88 | 568.18 | 813.74 | 1221.10 | 1255.45 |
| | 9 | 186.29 | 275.74 | 328.90 | 441.22 | 537.80 | 568.09 | 813.79 | 1221.05 | 1255.43 |
| | 10 | 186.38 | 275.79 | 328.88 | 441.16 | 537.86 | 568.03 | 813.82 | 1221.11 | 1255.49 |
| Avg(px) | | 186.34 | 275.77 | 328.92 | 441.26 | 537.77 | 568.14 | 813.79 | 1221.10 | 1255.45 |
| R | | 19.669 | 19.574 | 19.552 | 19.476 | 19.447 | 19.442 | 19.415 | 19.382 | 19.371 |

**Table B2**. Pixel-based diameter measurement results obtained for the metal set using the [7] method.

| | | Diameter Measurements [px] | | | | | | | | |
|---|---|---|---|---|---|---|---|---|---|---|
| | Actual Values | 3.665 mm | 5.398 mm | 6.431 mm | 8.594 mm | 10.458 mm | 11.046 mm | 15.8 mm | 23.668 mm | 24.32 mm |
| Measurements | 1 | 187.47 | 276.73 | 329.99 | 442.30 | 538.71 | 569.07 | 814.59 | 1221.85 | 1255.88 |
| | 2 | 187.52 | 276.70 | 329.88 | 442.20 | 538.65 | 568.91 | 814.61 | 1221.78 | 1256.08 |
| | 3 | 187.49 | 276.80 | 329.85 | 442.18 | 538.66 | 568.97 | 814.53 | 1221.84 | 1256.02 |
| | 4 | 187.43 | 276.86 | 329.77 | 442.20 | 538.52 | 568.95 | 814.47 | 1221.83 | 1255.83 |
| | 5 | 187.47 | 277.00 | 329.88 | 442.16 | 538.56 | 568.75 | 814.61 | 1221.85 | 1256.06 |
| | 6 | 187.47 | 276.76 | 329.78 | 442.28 | 538.66 | 569.10 | 814.45 | 1221.85 | 1255.97 |
| | 7 | 187.55 | 276.76 | 329.86 | 442.14 | 538.77 | 569.15 | 814.57 | 1221.82 | 1255.97 |
| | 8 | 187.49 | 276.63 | 329.82 | 442.22 | 538.73 | 569.03 | 814.44 | 1221.83 | 1255.99 |
| | 9 | 187.46 | 276.85 | 329.85 | 442.11 | 538.69 | 568.96 | 814.51 | 1221.80 | 1256.02 |
| | 10 | 187.53 | 276.95 | 329.79 | 442.03 | 538.77 | 568.79 | 814.63 | 1221.84 | 1256.08 |
| Avg(px) | | 187.49 | 276.81 | 329.85 | 442.18 | 538.67 | 568.97 | 814.54 | 1221.83 | 1255.99 |
| R | | 19.548 | 19.501 | 19.497 | 19.435 | 19.414 | 19.414 | 19.397 | 19.371 | 19.363 |

**Table B3**. Pixel-based diameter measurement results obtained for the metal set using the [33] method.